\documentclass{article}
\usepackage{spconf,amsmath,graphicx}
\usepackage{algorithm}
\usepackage{algorithmic}
\usepackage{booktabs}
\usepackage{tikz}
\usepackage{comment}
\usepackage{amsmath,amssymb} 
\usepackage{color}
\usepackage{multirow}
\usepackage{makecell}

\title{ChangeNet: Multi-Temporal Asymmetric  Change Detection Dataset}

\name{Deyi Ji$^{1, 2}$ \qquad Siqi Gao$^{2}$ \qquad Mingyuan Tao$^{2}$ \qquad Hongtao Lu$^{3, \star}$ \qquad  Feng Zhao$^{1, \star}$ \thanks{$^{\star}$Corresponding Authors.} \thanks{This work was supported by the National Key R\&D Program of China under Grant 2020AAA0103902, the Anhui Provincial Natural Science Foundation under Grant 2108085UD12, the JKW Research Funds under Grant 20-163-14-LZ-001-004-01, NSFC (No. 62176155), Shanghai Municipal Science and Technology Major Project, China (2021SHZDZX0102). We acknowledge the support of GPU cluster built by MCC Lab of Information Science and Technology Institution, USTC.}}

\address{$^{1}$University of Science and Technology of China \qquad  $^{2}$Alibaba Group\\ $^{3}$Dept. of CSE, MOE Key Lab of Artificial Intelligence, AI Institute, Shanghai Jiao Tong University}
\begin{document}
%
\maketitle
\begin{abstract}
Change Detection (CD) has been attracting extensive interests with the availability of bi-temporal datasets. However, due to the huge cost of multi-temporal images acquisition and labeling, existing change detection datasets are small in quantity, short in temporal, and low in practicability. Therefore, a large-scale practical-oriented dataset covering wide temporal phases is urgently needed to facilitate the community. To this end, the ChangeNet dataset is presented especially for multi-temporal change detection, along with the new task of ``Asymmetric Change Detection". Specifically, ChangeNet consists of 31,000 multi-temporal images pairs, a wide range of complex scenes from 100 cities, and  6 pixel-level  annotated categories, which is far superior to all the existing change detection datasets including LEVIR-CD, WHU Building CD, etc.. In addition, ChangeNet contains amounts of real-world perspective distortions in different temporal phases on the same areas, which is able to  promote the practical application of change detection algorithms. The ChangeNet dataset is suitable for both binary change detection (BCD) and semantic change detection (SCD) tasks. Accordingly, we benchmark the ChangeNet dataset  on six BCD methods and two SCD methods, and extensive experiments demonstrate its challenges and great significance. The dataset is available at https://github.com/jankyee/ChangeNet.
\end{abstract}
\section{Introduction}
\label{sec:intro}

Change Detection (CD) aims to provide the accurate object change information of land surface in bi-temporal or multi-temporal  remote sensing imagery, and is a meaningful and challenging fundamental task in various downstream remote sensing and earth vision area \cite{changestar, asymmetric}. 

According to the output types, change detection tasks can be divided into binary change detection (BCD) and semantic change detection.  Given a pair of bi-temporal images, BCD only focuses on addressing the locations of land-cover changed pixels between the input images, but fails to depict the semantic change information that is highly demanded in subsequent applications, since it overlooks the categories of pixels \cite{asymmetric}. Therefore, SCD is subsequently proposed to simultaneously identify both the change regions and their land-cover categories in bi-temporal images. 

Different from many other tasks that only require single temporal images,  bi-temporal or even multi-temporal imagery with long time spans are essential to the performance of change detection methods. Correspondingly, there are several commonly-used datasets in existing change detection methods, including LEVIR-CD \cite{LEVIR-CD}, WHU building CD \cite{WHU}, DSIFN-CD \cite{DSIFN-CD}, and SECOND \cite{asymmetric}. The former three are for BCD while the latter one is for SCD. However, there are three shortcomings in the existing datasets: 

\begin{figure}
    \centering
    \includegraphics[width=1\linewidth]{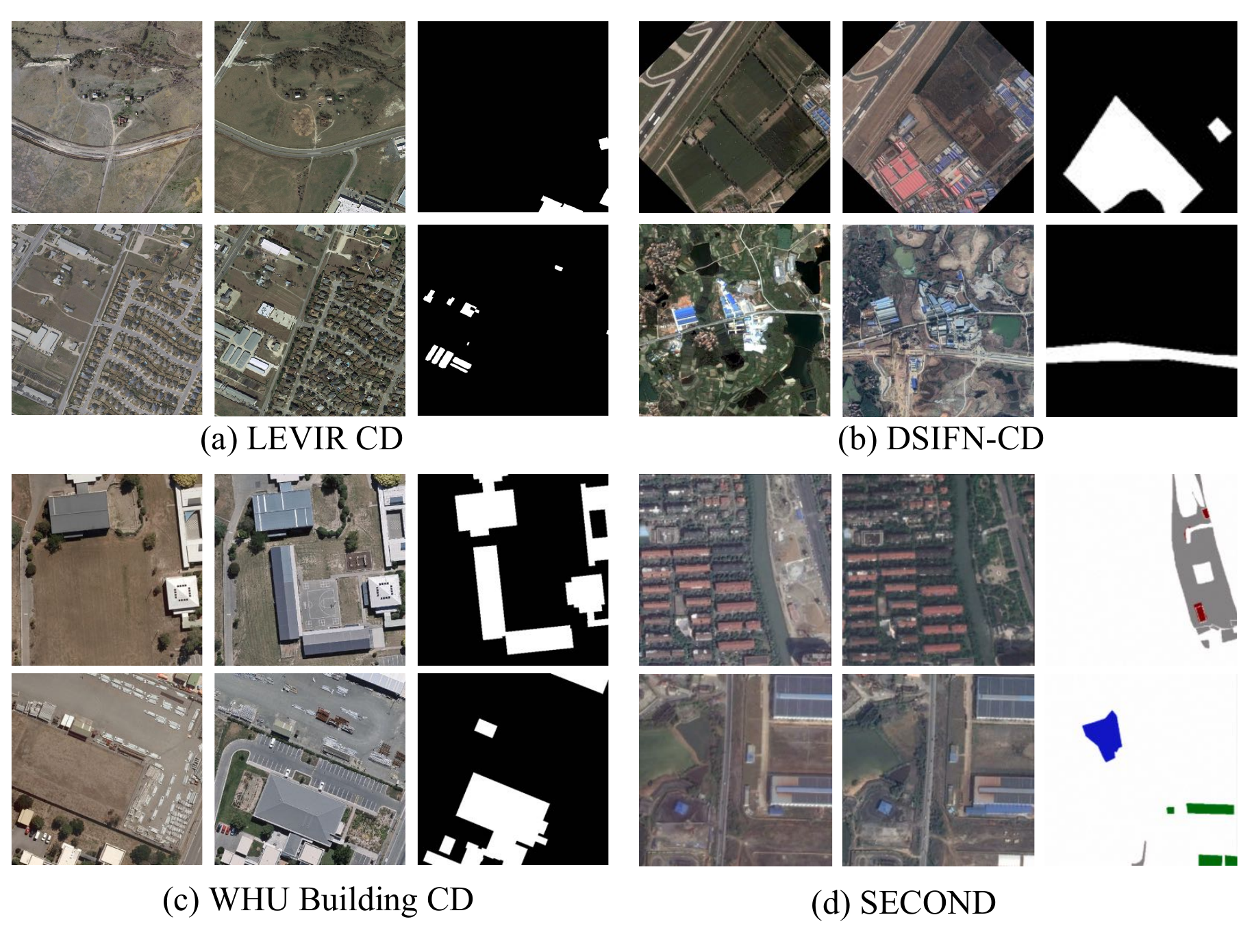}
    \caption{The existing bi-temporal change detection datasets, including LEVIR-CD \cite{LEVIR-CD}, DSIFN-CD \cite{DSIFN-CD}, WHU Building CD \cite{WHU}, SECOND \cite{asymmetric}. The former three are only for binary change detection (BCD), while the latter is applicable to semantic change detection.  Existing change detection datasets usually consist of short (only two) temporal phases with completely symmetrical land-cover appearances.}
    \label{intro1}
\end{figure}

\begin{itemize}
    \item Small Quantity: Reflected in both images number and resolution.  WHU building CD and LEVIR-CD only contain 2 and 637 images respectively. Although DSIFN-CD and SECOND consist of more images (3,988 and 2,968 respectively), their resolutions are only 512$\times$512.
    
    \item Short Temporal: There are almost only two temporal phases in existing datasets. However, multi-temporal data with a long time span is crucial to characterize the land-cover change in a region.
    
    \item Low Practicability: In the existing datasets, all images in different temporal phases are captured under fixed parameters, thus are basically consistent in appearance without any distortion. However, in practice, there are differences in equipment parameters in years cross-temporal  collection, so the collected data also has different distortions, which brings great challenges to the algorithm to be solved in 
    practical applications. In this paper, we define this practical problem as ``Asymmetric Change Detection".
\end{itemize}

Specifically, as shown in Figure \ref{intro1}, in the existing change detection datasets, images in different temporal phases are strictly symmetrical at the pixel level for the convenience of algorithm development, that is, the spatial coordinates of the change area in images of different temporal phases are the same. However, in actual remote sensing acquisition, due to some inevitable differences in acquisition equipment, weather, viewing angle and altitude in each phase of acquisition, the collected images can not guarantee strict pixel-level alignment on the ground area, and there may be serious viewing angle and apparent shape distortion. Hence, it is not sufficient to fully verify the performance of change detection methods with existing datasets, which limits the development of the community.

\begin{figure}
    \centering
    \includegraphics[width=1\linewidth]{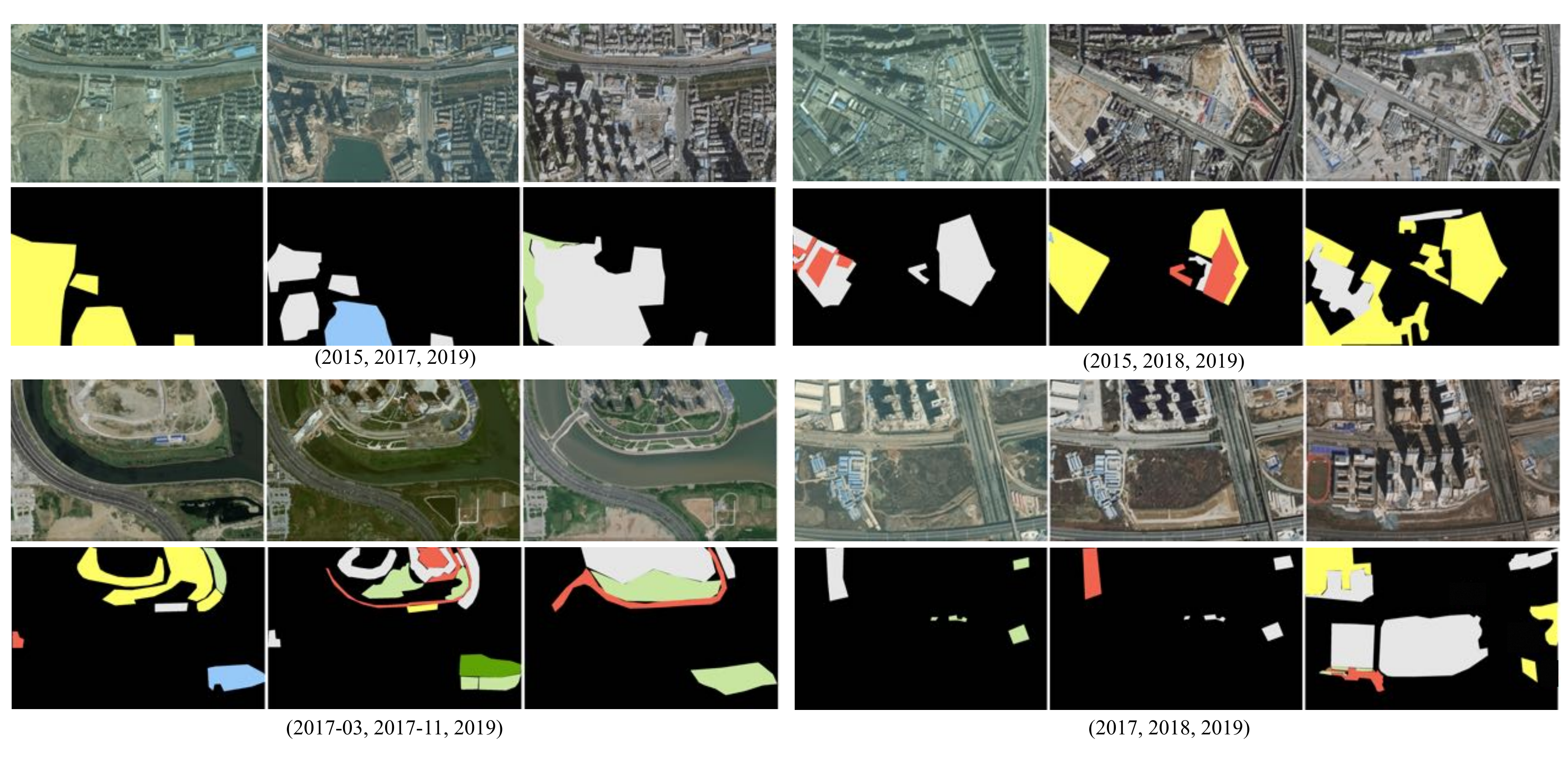}
    \caption{The ChangeNet contains images covering a wide range of time phases and the land-cover change annotations. There are up to 6 temporal phases with amounts of practical asymmetric changes. Here, we select four samples with 3 temporal phases.}
    \label{intro}
\end{figure}

Therefore, a novel large-scale, practical-oriented multi-temporal dataset for asymmetric change detection with dense annotations is urgently needed to facilitate the field and promote the practicality of change detection methods. To this end, the ChangeNet datasets is proposed. First of all, in terms of quantity, the ChangeNet dataset has 31,000 multi-temporal images, which is much larger than all existing change detection datasets. Secondly,  the temporal  of the ChangeNet dataset is up to six phases (from 2014 to 2022), which are also longer than other existing change detection datasets. Last but not least, the scenes of ChangeNet are more complex and practical, including a large number of real-world changes with different perspective distortions.  The ChangeNet dataset is suitable for both binary change detection (BCD) and semantic change detection (SCD) tasks, and able to  promote the practical application of change detection algorithms.

\section{Related Work}

Compared to single-temporal datasets, multi-temporal datasets are more difficult and expensive to acquire and annotate, so they are less involved in existing research works \cite{ipgn,cagcn,ji2019end,feng2018challenges}. Most of existing change detection methods are inspired from image recognition and segmentation methods \cite{stlnet,urur,cdgc,gpwformer,sstkd,zhu2023llafs}. Generally, multi-temporal datasets are mostly used in land resources management and other related fields and have driven the research of land resources evolution, such as urbanization, land desertification, etc. Existing change detection datasets are usually divided into binary change detection and semantic change detection according to whether there are multi-semantic category annotations. The former means that it only needs to distinguish whether there is a change in a certain area in multiple time phases, and the representative datasets are LEVIR-CD, WHU Building CD, and DFIFN-CD, while the latter not only needs to distinguish whether it has changes, but also needs to distinguish from which category to which category, the representative dataset is SECOND. Obviously, semantic change detection datasets are more difficult in actual data acquisition and labeling, so the current datasets are often small in size, low in resolution, or sparse in categories.

\section{ChangeNet Dataset}

\subsection{Overview}

The proposed ChangeNet dataset is far superior to all the existing Change Detection datasets including WHU Budiling CD \cite{WHU}, LEVIR-CD \cite{LEVIR-CD}, DSIFN-CD \cite{DSIFN-CD}, Hi-UCD \cite{hi-ucd} and SECOND \cite{asymmetric}, in both quantity and practicability. Specifically, ChangeNet contains 31,000 images with resolution of 0.3 meters and size of 1,900 $\times$ 1,200, from 100 cities in China, where about 10,000 images contain obvious changes.  The training, validation and testing set include 21,700, 3,100 and 6,200 images respectively, with the approximate ratio of 7:1:2. The images are manually annotated with fine-grained pixel-level categories, including 6 classes of “building”, “farmland”, “bareland”, “water”, “road” and “unchanged”. Sample images are shown in Figure \ref{intro}.

\begin{table}[]
\centering
\caption{The detailed statistics comparison between ChangeNet and the existing change detection datasets. As shown that ChangeNet is far superior to all of them in terms of both quantity and practicability. Note that Hi-UCD is not publicly available, and SECOND only releases 2968 images.}
\scalebox{0.72}{
\begin{tabular}{c|ccccccc}
\toprule
 \textbf{Dataset}  & \multirow{2}{*}{Image} & \multirow{2}{*}{\begin{tabular}[c]{@{}c@{}} Resolution \\ (m) \end{tabular}}  & \multirow{2}{*}{Class} & \multirow{2}{*}{Temporal} & \multirow{2}{*}{City} & \multicolumn{2}{c}{Task} \\ \cmidrule{7-8} 
                                          &                         &                                                              &                             &                           &                         & BCD         & SCD         \\ \midrule
WHU-CD                           & 2                       & 0.3                                              & 2                           & 2            & 1                       &      $\checkmark$       &             \\
LEVIR-CD                                  & 637                     & 0.5                                                & 2                           & 2                         & -                       &    $\checkmark$         &             \\
DSIFN-CD                                  & 3988                    & 0.5                                                  & 2                           & 2                         & 6                       &      $\checkmark$       &             \\
Hi-UCD$^*$                                & 745                     & 0.1
                       & 9                       & 2
                & -                       &                 & $\checkmark$ \\
SECOND                                    & 2968                    & 0.3$\sim$5                                                  & 6                           & 2                         & 3                       & $\checkmark$            &   $\checkmark$          \\ \midrule
ChangeNet                                 & 31000                   & 0.3                                          & 6                           & 6                         & 100                     & $\checkmark$ & $\checkmark$             \\ \bottomrule
\end{tabular}}

\end{table}

\subsection{Dataset Collection}

The ChangeNet dataset is constructed using 0.3m images captured from the WayBack platform \cite{wayback} from 2014 to 2022 for public use.
This platform provides rich interfaces for obtaining multi-temporal data. 
When collecting the data of 100 cities, each annotator is responsible for the collection of 10 cities with  randomly selected 100 non-overlapping areas. After the very first  pre-processing, we totally obtain 31,805 image sets, with an average of four images per image set.

\subsection{Efficient Annotation}

Compared to single-temporal images, annotating the multi-temporal images sets is always a more tough job, as multi-temporal not only means a rapid increase in the number of targets to be annotated, but also leads to heavy regional comparison work between different images. In addition, due to the different shooting settings, there may be serious perspective distortions in images under different phases in a  ground area, which also brings the challenge to the multi-temporal annotation. Based on the above analysis, in order to balance the  quality and cost, we utilize an intuitive and effective annotation scheme. Considering a  multi-temporal image set $I$ to be annotated, $I_i$ $(i \in [1, N])$ denotes the $i$th temporal sub-image where $N$ is the total number of temporal. Since the entire image set contains rich ground elements and context information, considering the convenience for annotators, we only display two images of different phases on the annotation panel. Hence the most direct way to annotate the change relationship between pairs of $N$ images is to annotate them in pairs, so that for each set of images, a total of $\frac{N \times (N-1)}{2}$ annotations are required. This scheme is direct yet resulting in heavy amounts of annotations. However, in practice, most of the changes of ground objects across multiple temporal phases change linearly with time, for example, in the process of urbanization of a city, the changes of buildings generally increase with the passage of time, and the process of land desertification is generally the gradual withering of vegetation over time. Based on this observation, we design an incremental annotation strategy, that is, for $N$ images arranged in chronological order, we only combine and annotate the following images pairs $\{I_i, I_j\}$ $(i<j\in[1, N])$. In this way, we greatly reduce the labeling burden. Moreover, for more efficient annotation, we co-develop a model with model-in-the-loop dataset annotation, the model is used to generate reference masks for the annotators to save manpower. As a reference, annotators adjust the masks with the help of annotation tools developed by us \cite{wang2021learning}.

\subsection{Dataset Statistics}

Table 1 shows the detailed statistics comparison between the proposed ChangeNet dataset and exiting several main change detection datasets, including WHU Building CD, LEVIR-CD, DSIFN-CD, Hi-UCD and SECOND. The former three are for binary change detection, while the latter two are for semantic change detection. Firstly, for quantity, ChangeNet consists of 31,000 images with average size of 1900$\times$1200, while the numbers of images in WHU Building CD, LEVID-CD and Hi-UCD are below 1,000. Despite that DSIFN-CD and SECOND consists of more images up to about 4,000, their image sizes are very low (512$\times$512). The spatial resolution of ChangeNet reaches 0.3m, which is also accurate enough. As for the total temporal phases, ChangeNet contains absolute advantage of 6 phases, which is far superior to all other datasets that only contains 2 temporal phases.  For  annotation, limited by manpower, the existing datasets  only annotate either binary change masks or small amount of images. The former indicates the binary change detection datasets including WHU Building CD, LEVIR-CD, and DSIFN-CD, and the latter indicates the semantic change detection datasets including Hi-UCD and SECOND.  By contrast, ChangeNet is  annotated with amounts of dense pixels of five categories change masks.  As a result, the ChangeNet dataset is appropriate to both binary and semantic change detection tasks. Moreover, ChangeNet contains amounts of real-world perspective distortions in different temporal phases on the same areas, thus is able to promote the practical application of change detection algorithms, based on which we further put forward the multi-temporal asymmetric change detection benchmark.

\section{Asymmetric Change Detection Benchmark}

For the first time, we present the task of ``Asymmetric Change Detection" with the ChangeNet dataset, which is applicable for both binary and semantic change detection. In this section, we base on previous methods and formulate the baselines for both binary asymmetric change detection (BACD) and semantic asymmetric change detection (SACD). 

\subsection{Binary Asymmetric Change Detection}

As BCD is the fundamental CD problem, we firstly build the benchmark for BACD.

\subsubsection{Experimental Settings}

To prove the significance of ChangeNet dataset, we validate several representative BCD methods on ChangNet, LEVIR-CD, DSIFN-CD for comparison. 
For fair and clear comparison, in our experiments we only re-implement and validate the performances of open-sourced state-of-the-art methods on ChangeNet dataset. We follow the previous works \cite{changeformer} to select the above  representative BCD methods, and follow their default training and inference setting. When performing experiments on BCD, the networks need only to predict the changed area, without any semantic categories. 

\subsubsection{Results and Discussions}

Table \ref{exp_bacd} shows the BACD benchmark, specifically, the comparison of performance of  six representative BCD methods on LEVIR-CD, DFIFN-CD and ChangeNet datasets. Generally, the two advanced methods ChangeSTAR and ChangeFormer show remarkable results on both the three datasets. Besides, the performances on  ChangeNet are obviously much lower than the ones on LEVIR-CD and DSIFN-CD, which demonstrate the greater challenges of ChangeNet. 
To further prove the significance of ChangeNet, we also conduct experiments to shown its generality by applying its pretraining experiments on LEVIR-CD, in Table \ref{exp_bacd_pretrain}, where the default pretraining is on ImageNet. Results show that ChangeNet pretraining achieves higher performances, proving its excellent potential. In conclusion, both the experiments demonstrate the significance of ChangeNet to advance the further research of change detection.

\begin{table}[]
\centering
\caption{The BACD benchmark. }
\scalebox{0.7}{\begin{tabular}{c|cc|cc|cc}
\toprule
\multirow{2}{*}{\textbf{Method}} & \multicolumn{2}{c|}{LEVIR-CD} & \multicolumn{2}{c|}{DSIFN-CD} & \multicolumn{2}{c}{ChangeNet} \\ \cmidrule{2-7} 
                         & IoU (\%)            & F1 (\%)             & IoU  (\%)          & F1 (\%)             & IoU (\%)            & F1  (\%)           \\ \midrule
IFNet \cite{DSIFN-CD}                   & 78.8          & 88.1          & 42.9          & 60.1          & 31.2          & 39.9          \\
SNUNet \cite{DSIFN-CD}                  & 78.8          & 88.2          & 49.5          & 66.2          & 30.8          & 39.3          \\
BIT  \cite{BIT}                    & 80.7          & 89.3          & 53.0          & 69.3          & 31.4          & 39.8          \\
ChangeSTAR \cite{changestar}              & 83.2          & 90.8          & -             & -             & 32.4          & 41.1          \\
ChangeFormer \cite{changeformer}            & 82.5          & 90.4          & 76.5          & 86.7          & 32.5          & 40.8          \\ \bottomrule
\end{tabular}}
\label{exp_bacd}
\end{table}

\subsection{Semantic Asymmetric Change Detection}

Compared to BACD, SACD is more informative to provide both binary change maps and their semantic categories indicating ``from-to" change direction, thus more challenging.

\subsubsection{Experimental Settings}

Compared to BCD datasets, the number of SCD datasets are relative much smaller, due to the expensive collection and annotation costs. Existing representative SCD datasets include Hi-UCD \cite{hi-ucd} and SECOND \cite{asymmetric}. Since Hi-UCD is not publicly available now, we mainly conduct the experiments on the SECOND dataset. We firstly describe it in detail as follows.

Unfortunately, due to the great challenge of SCD, there are few related studies and datasets and unified frameworks are still not well addressed. In our experiments, we also select two open-sourced frameworks, including ChangeMask \cite{changemask} and  HRSCD \cite{DSIFN-CD}. We also follow their default  setting in our experiments.

\begin{table}[]
\caption{The effectiveness of ChangeNet pretraining.}
\centering
\scalebox{0.8}{\begin{tabular}{c|c|c}
\toprule
\textbf{Method}       & ChangeSTAR IoU(\%) & ChangeFormer IoU(\%) \\ \midrule
Default Pretraning    &  83.2           & 82.5                       \\
ChangeNet Pretraining &   84.9         &   85.1                    \\ \bottomrule
\end{tabular}}
\label{exp_bacd_pretrain}
\end{table}

\begin{table}[]
\caption{The SACD benchmark. }
\centering
\scalebox{0.9}{\begin{tabular}{c|cc|cc}
\toprule
\multirow{2}{*}{\textbf{Method}} & \multicolumn{2}{c|}{SECOND} & \multicolumn{2}{c}{ChangeNet} \\ \cmidrule{2-5} 
                                 & IoU (\%)       & F1 (\%)      & IoU (\%)        & F1 (\%)       \\ \midrule
HRSCD \cite{DSIFN-CD}                           & 42.0          & 59.2        & 20.7           & 33.2         \\
ChangeMask \cite{changemask}                      & 54.3          & 70.3        & 24.4           & 36.9         \\ \bottomrule
\end{tabular}}
\label{exp_sacd}
\end{table}

\subsubsection{Results and Discussions}

Table \ref{exp_sacd} shows the SACD benchmark, with the performances of two representative SCD methods on SECOND and ChangeNet datasets.
Notably, the results obtained on the ChangeNet dataset are noticeably inferior to those on SECOND, which not only underscores the formidable challenge posed by ChangeNet but also its considerable importance in pushing the boundaries of current change detection capabilities.

\section{Conclusion}

In this paper, we firstly analyze the shortcomings of the existing change detection datasets: small quantity, short temporal and low practicability. 
To this end, the ChangeNet dataset is proposed especially for multi-temporal change detection, along with the new task of ``Asymmetric Change Detection". ChangeNet is  far superior to all the existing change detection datasets, and contains amounts of real-world perspective distortions in different temporal phases on the same areas, which is able to promote the practical application of change detection algorithms. 
We present the BACD and SACD tasks, and show that both of them are applicable to the ChangeNet dataset. Finally,
extensive benchmark experiments on several change detection methods demonstrate the challenges and great significance of the ChangeNet dataset.

\vfill\pagebreak

\bibliographystyle{IEEEbib}
\bibliography{strings,refs}

\begin{thebibliography}{10}

\bibitem{changestar}
Zhuo Zheng, Ailong Ma, Liangpei Zhang, and Yanfei Zhong,
\newblock ``Change is everywhere: Single-temporal supervised object change detection in remote sensing imagery,''
\newblock in {\em IEEE/CVF International Conference on Computer Vision}, 2021, pp. 15193--15202.

\bibitem{asymmetric}
Kunping Yang, Gui-Song Xia, Zicheng Liu, Bo~Du, Wen Yang, Marcello Pelillo, and Liangpei Zhang,
\newblock ``Asymmetric siamese networks for semantic change detection in aerial images,''
\newblock {\em IEEE Transactions on Geoscience and Remote Sensing}, vol. 60, pp. 1--18, 2021.

\bibitem{LEVIR-CD}
Hao Chen and Zhenwei Shi,
\newblock ``A spatial-temporal attention-based method and a new dataset for remote sensing image change detection,''
\newblock {\em Remote Sensing}, vol. 12, no. 10, pp. 1662, 2020.

\bibitem{WHU}
Shunping Ji, Shiqing Wei, and Meng Lu,
\newblock ``Fully convolutional networks for multisource building extraction from an open aerial and satellite imagery data set,''
\newblock {\em IEEE Transactions on Geoscience and Remote Sensing}, vol. 57, no. 1, pp. 574--586, 2018.

\bibitem{DSIFN-CD}
Chenxiao Zhang, Peng Yue, Deodato Tapete, Liangcun Jiang, Boyi Shangguan, Li~Huang, and Guangchao Liu,
\newblock ``A deeply supervised image fusion network for change detection in high resolution bi-temporal remote sensing images,''
\newblock {\em ISPRS Journal of Photogrammetry and Remote Sensing}, vol. 166, pp. 183--200, 2020.

\bibitem{ipgn}
Haoran Wang, Licheng Jiao, Fang Liu, Lingling Li, Xu~Liu, Deyi Ji, and Weihao Gan,
\newblock ``Ipgn: Interactiveness proposal graph network for human-object interaction detection,''
\newblock {\em IEEE Transactions on Image Processing}, vol. 30, pp. 6583--6593, 2021.

\bibitem{cagcn}
Deyi Ji, Haoran Wang, Hanzhe Hu, Weihao Gan, Wei Wu, and Junjie Yan,
\newblock ``Context-aware graph convolution network for target re-identification,''
\newblock {\em arXiv preprint arXiv:2012.04298}, 2020.

\bibitem{ji2019end}
Deyi Ji, Hongtao Lu, and Tongzhen Zhang,
\newblock ``End to end multi-scale convolutional neural network for crowd counting,''
\newblock in {\em Eleventh international conference on machine vision}. SPIE, 2019, vol. 11041, pp. 761--766.

\bibitem{feng2018challenges}
Weitao Feng, Deyi Ji, Yiru Wang, Shuorong Chang, Hansheng Ren, and Weihao Gan,
\newblock ``Challenges on large scale surveillance video analysis,''
\newblock in {\em IEEE Conference on Computer Vision and Pattern Recognition Workshops}, 2018, pp. 69--76.

\bibitem{stlnet}
Lanyun Zhu, Deyi Ji, Shiping Zhu, Weihao Gan, Wei Wu, and Junjie Yan,
\newblock ``Learning statistical texture for semantic segmentation,''
\newblock in {\em IEEE/CVF Conference on Computer Vision and Pattern Recognition}, June 2021, pp. 12537--12546.

\bibitem{urur}
Deyi Ji, Feng Zhao, Hongtao Lu, Mingyuan Tao, and Jieping Ye,
\newblock ``Ultra-high resolution segmentation with ultra-rich context: A novel benchmark,''
\newblock in {\em IEEE/CVF Conference on Computer Vision and Pattern Recognition}, 2023, pp. 23621--23630.

\bibitem{cdgc}
Hanzhe Hu, Deyi Ji, Weihao Gan, Shuai Bai, Wei Wu, and Junjie Yan,
\newblock ``Class-wise dynamic graph convolution for semantic segmentation,''
\newblock in {\em European Conference on Computer Vision}. Springer, 2020, pp. 1--17.

\bibitem{gpwformer}
Deyi Ji, Feng Zhao, and Hongtao Lu,
\newblock ``Guided patch-grouping wavelet transformer with spatial congruence for ultra-high resolution segmentation,''
\newblock {\em International Joint Conference on Artificial Intelligence}, 2023.

\bibitem{sstkd}
Deyi Ji, Haoran Wang, Mingyuan Tao, Jianqiang Huang, Xian-Sheng Hua, and Hongtao Lu,
\newblock ``Structural and statistical texture knowledge distillation for semantic segmentation,''
\newblock in {\em IEEE/CVF Conference on Computer Vision and Pattern Recognition}, 2022, pp. 16876--16885.

\bibitem{zhu2023llafs}
Lanyun Zhu, Tianrun Chen, Deyi Ji, Jieping Ye, and Jun Liu,
\newblock ``Llafs: When large-language models meet few-shot segmentation,''
\newblock {\em arXiv preprint arXiv:2311.16926}, 2023.

\bibitem{hi-ucd}
Shiqi Tian, Ailong Ma, Zhuo Zheng, and Yanfei Zhong,
\newblock ``Hi-ucd: A large-scale dataset for urban semantic change detection in remote sensing imagery,''
\newblock {\em arXiv preprint arXiv:2011.03247}, 2020.

\bibitem{wayback}
``World imagery wayback,''
\newblock {\em Availabe online: https://livingatlas.arcgis.com/wayback/}, 2023.

\bibitem{wang2021learning}
Haoran Wang, Licheng Jiao, Fang Liu, Lingling Li, Xu~Liu, Deyi Ji, and Weihao Gan,
\newblock ``Learning social spatio-temporal relation graph in the wild and a video benchmark,''
\newblock {\em IEEE Transactions on Neural Networks and Learning Systems}, 2021.

\bibitem{changeformer}
Wele Gedara~Chaminda Bandara and Vishal~M Patel,
\newblock ``A transformer-based siamese network for change detection,''
\newblock in {\em IEEE International Geoscience and Remote Sensing Symposium}. IEEE, 2022, pp. 207--210.

\bibitem{BIT}
Hao Chen, Zipeng Qi, and Zhenwei Shi,
\newblock ``Remote sensing image change detection with transformers,''
\newblock {\em IEEE Transactions on Geoscience and Remote Sensing}, vol. 60, pp. 1--14, 2021.

\bibitem{changemask}
Zhuo Zheng, Yanfei Zhong, Shiqi Tian, Ailong Ma, and Liangpei Zhang,
\newblock ``Changemask: Deep multi-task encoder-transformer-decoder architecture for semantic change detection,''
\newblock {\em ISPRS Journal of Photogrammetry and Remote Sensing}, vol. 183, pp. 228--239, 2022.

\end{thebibliography}

\end{document}